%% file: main.tex
\documentclass[conference]{IEEEtran}
\IEEEoverridecommandlockouts
\usepackage{cite}
\usepackage{amsmath,amssymb,amsfonts}
\usepackage{graphicx}
\usepackage{textcomp}
\usepackage[whole]{bxcjkjatype} % for Japanese
\usepackage{multirow}
\usepackage{subcaption}
\usepackage[table,xcdraw]{xcolor}
\usepackage{threeparttable}
\usepackage{eucal}
\usepackage{balance}
\usepackage{mathrsfs}  
\usepackage{comment}
\usepackage{algorithm}
\usepackage[noend]{algpseudocode}
\usepackage{float}
\usepackage{textcomp}
\usepackage{mathcomp}
\usepackage{bm}
\usepackage{cite}
\usepackage{url}
\usepackage{flushend} % 最終ページの下側調整
\usepackage[normalem]{ulem}
\usepackage{makecell}

\def\ie{\textit{i.e}\onedot}

\def\etal{\textit{et al}\onedot}

\newcommand{\myvec}[1]{\mathbf{#1}}

\newcommand{\myset}[1]{\mathcal{#1}}

\def\ie{{\it i.e.}}
\def\etal{{\it et al. }}

\DeclareRobustCommand{\erase}{\bgroup\markoverwith{\textcolor{red}{\rule[.5ex]{2pt}{0.4pt}}}\ULon}
\newcommand\redout{\bgroup\markoverwith
{\textcolor{red}{\rule[.5ex]{2pt}{0.4pt}}}\ULon}
\def\BibTeX{{\rm B\kern-.05em{\sc i\kern-.025em b}\kern-.08em
    T\kern-.1667em\lower.7ex\hbox{E}\kern-.125emX}}
\begin{document}

\title{
LVLM-MPC Collaboration for Autonomous Driving: A Safety-Aware and Task-Scalable Control Architecture
}

\author{
\IEEEauthorblockN{1\textsuperscript{st} Kazuki Atsuta}
\IEEEauthorblockA{\textit{Department of Mechanical Systems Engineering} \\
\textit{Nagoya University}\\
Nagoya, Japan \\
atsuta.kazuki.j6@s.mail.nagoya-u.ac.jp}
\and
\IEEEauthorblockN{2\textsuperscript{nd} Kohei Honda}
\IEEEauthorblockA{\textit{Department of Mechanical Systems Engineering} \\
\textit{Nagoya University}\\
Nagoya, Japan \\
honda.kohei.k3@f.mail.nagoya-u.ac.jp}
\and
\IEEEauthorblockN{3\textsuperscript{rd} Hiroyuki Okuda}
\IEEEauthorblockA{\textit{Department of Mechanical Systems Engineering} \\
\textit{Nagoya University}\\
Nagoya, Japan \\
h\_okuda@nuem.nagoya-u.ac.jp}
\and
\IEEEauthorblockN{4\textsuperscript{th} Tatsuya Suzuki}
\IEEEauthorblockA{\textit{Department of Mechanical Systems Engineering} \\
\textit{Nagoya University}\\
Nagoya, Japan \\
t\_suzuki@nuem.nagoya-u.ac.jp}
% \and
% \and
}

\maketitle

\input{src/abstract}


\input{src/introduction}
\input{src/related_work}
\input{src/preliminary}

\input{src/method}

\input{src/experiment}
\input{src/conclusion}

\bibliographystyle{IEEEtran}
\bibliography{IEEEabrv, reference}

\end{document}

%% file: src/abstract.tex
\begin{abstract}
This paper proposes a novel Large Vision-Language Model (LVLM) and Model Predictive Control (MPC) integration framework that delivers both task scalability and safety for Autonomous Driving (AD).
LVLMs excel at high-level task planning across diverse driving scenarios.
However, since these foundation models are not specifically designed for driving and their reasoning is not consistent with the feasibility of low-level motion planning, concerns remain regarding safety and smooth task switching.
This paper integrates LVLMs with MPC Builder, which automatically generates MPCs on demand, based on symbolic task commands generated by the LVLM, while ensuring optimality and safety.
The generated MPCs can strongly assist the execution or rejection of LVLM-driven task switching by providing feedback on the feasibility of the given tasks and generating task-switching-aware MPCs.
Our approach provides a safe, flexible, and adaptable control framework, bridging the gap between cutting-edge foundation models and reliable vehicle operation.
We demonstrate the effectiveness of our approach through a simulation experiment, showing that our system can safely and effectively handle highway driving while maintaining the flexibility and adaptability of LVLMs.
\end{abstract}

%% file: src/introduction.tex
\section{Introduction} \label{sec:introduction}
\begin{figure}[t]
  \centering
  \includegraphics[width=\linewidth]{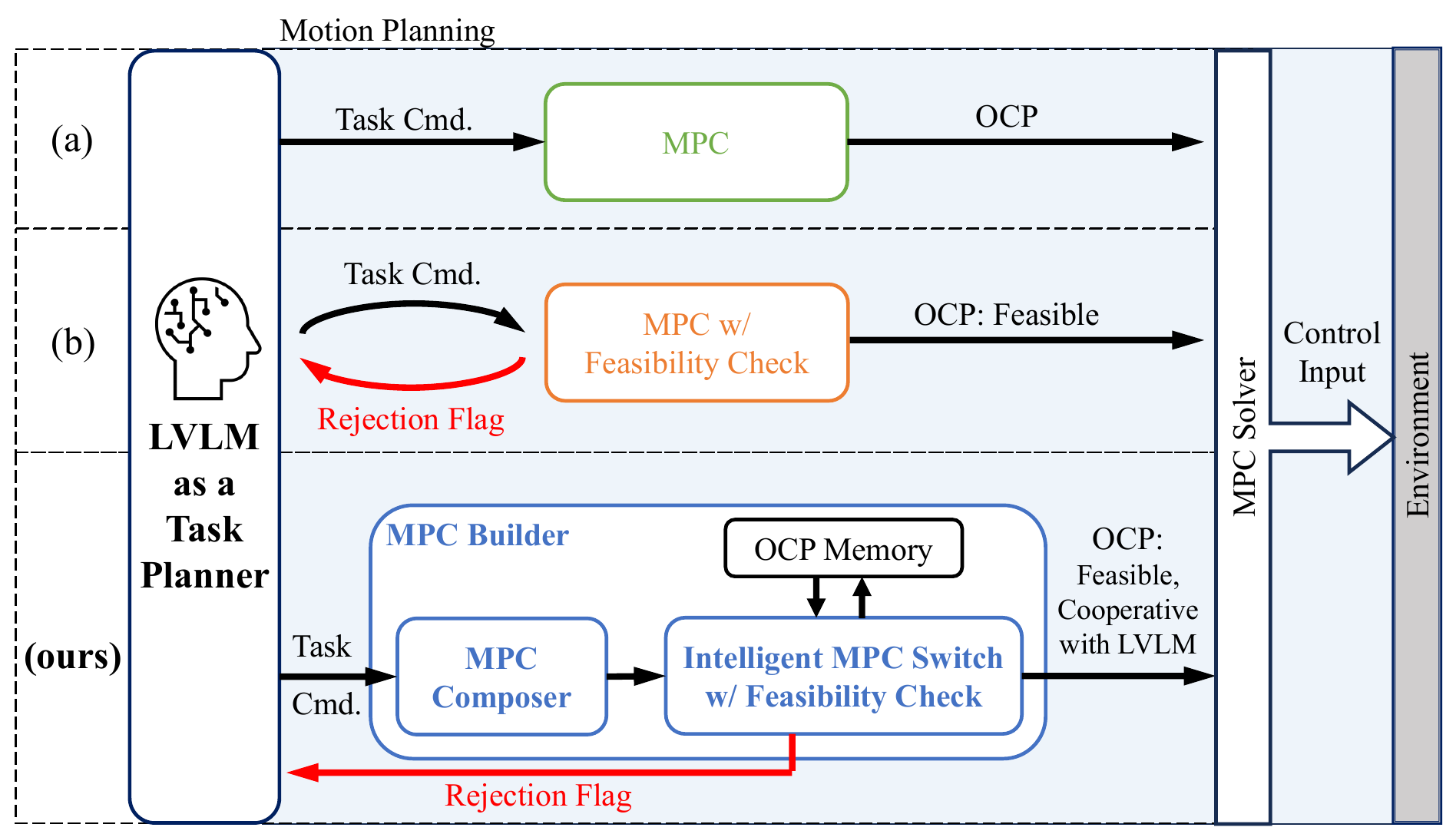}
  \caption{
  % Concept of the proposed framework.
  Conceptual comparison of hierarchical LVLM-MPC pipelines.
(a) A unidirectional LVLM-to-MPC chain either executes tasks or stalls; (b) a bidirectional variant merely returns a binary infeasibility flag; (ours) a feedback loop, powered by MPC Builder, assembles task-specific Optimal Control Problems (OCPs) and synthesizes safe recovery maneuvers, enabling seamless task–motion handovers.
  }
  \vspace{-1.5em}
  \label{fig:concept}
\end{figure}

Imagine a human driver weaving through heavy traffic on a congested road.  
They read overhead signs, anticipate merges and splits, and select an optimal lane well in advance, while fine-tuning steering, throttle, and brake inputs to perform various driving subtasks.  
This seamless blend of long-horizon task planning and real-time motion control is essential for autonomous driving~(AD).  
To bridge this semantic control gap, we propose a \emph{hierarchical LVLM-MPC architecture} in which a Large Vision-Language Model~(LVLM)~\cite{gpt-4o} produces symbolic task commands from visual and linguistic cues, and a Model Predictive Controller~(MPC) executes them under dynamic constraints, as illustrated in Fig.~\ref{fig:concept}.  

Although similar hierarchical frameworks have been studied, two key challenges remain unaddressed.  
The first challenge is the limited scalability of conventional model-based planners such as MPC when paired with the open-ended task space generated by modern LVLMs.  
Most model-based controllers are hand-crafted for a small set of tasks, which restricts their versatility.  
Expanding their capability typically requires either switching among multiple controllers~\cite{MPC_switch} or formulating a large multi-objective optimization problem~\cite{MPC_MixedInteger}.  
As the number of tasks grows, the design and verification effort increases exponentially, which is a severe drawback when combined with an LVLM that can output diverse instructions.  

The second challenge concerns communication when a requested task is temporarily infeasible.  
Many prior works adopt a unidirectional pipeline (Fig.~\ref{fig:concept}(a)): the LVLM issues a high-level command, and the MPC either succeeds or fails silently, compromising task--motion consistency~\cite{LanguageMPC,bidirectional}.  
Some methods let the MPC return a binary \emph{infeasible} flag (Fig.~\ref{fig:concept}(b))~\cite{ASafetyPerspective}.  
However, a lone flag can leave the vehicle trapped in a state where the LVLM's proposal remains infeasible, forcing repeated re-planning, deadlocks, and wasted computation.  

\looseness=-1
To address these challenges, we integrate LVLM-based planning with MPC-based execution through \emph{MPC Builder}~\cite{MPCBuilder}, as depicted in Fig.~\ref{fig:concept}.  
MPC Builder overcomes scalability by automatically synthesizing MPCs from a compact library of primitive design elements, enabling rapid adaptation to diverse environments and complex tasks.  
For infeasible tasks, it not only rejects the command but also constructs an alternative optimal control problem that drives the system toward a feasible state, thereby assisting smooth task switching.  
Consequently, MPC Builder both facilitates successive task handovers from the LVLM and guarantees feasibility and safety during execution.  

We validate the framework in highway-traffic simulations with multiple surrounding vehicles.  
Baseline methods in Fig.~\ref{fig:concept}(a) and (b) frequently violate safety constraints and suffer deadlocks.  
In contrast, our approach attains a 100\% success rate in safely navigating the scenario through tight LVLM--MPC cooperation.  
Moreover, whereas the baselines rely heavily on prompt engineering, our method maintains robust safety \emph{without} task-specific safety prompts, thanks to MPC Builder's inherent safe task-switching assistance.  

%% file: src/related_work.tex
\section{Related Work} \label{sec:related_work}

\subsection{Language-model-based Task and Motion Planning for AD}\label{ssec:LLM_tamp_AD}
AD must accomplish an enormous variety of driving tasks, and their probability distribution exhibits a long tail \cite{long-tailed}, meaning that many scenarios occur with low frequency.  
Therefore, achieving general-purpose AD is beyond the scope of scenario-specific methods such as rule-based methods \cite{RuleLaneChange}, imitation learning \cite{DLLaneChange}, or reinforcement learning approaches \cite{RLLaneChange}.  

To address this challenge, the recent dramatic performance gains of language models such as large language models~(LLMs) and LVLMs have spurred intensive research into their incorporation into autonomous driving \cite{LLM4Drive,MLLMforAD}.  
Thanks to their reasoning, interpretability, knowledge, and memory capabilities, LLMs and LVLMs are expected to handle long-tail corner cases at a human level \cite{DriveLikeAHuman}, drawing attention to their potential as versatile, closed-loop task planners for planning and decision making.  
Several studies propose end-to-end methods in which an LLM directly outputs waypoints or control inputs \cite{LMDrive,GPTDriver,DriveGPT4}.  
However, such methods can generate actions that violate vehicle kinematics or dynamics, thus failing to guarantee safety.  
Moreover, the response latency of large-scale models is far greater than the control cycle required for autonomous driving, posing a significant barrier to real-time deployment.  

To enhance safety and improve real-time performance, some studies adopt a two-layer approach in which an LLM serves as a high-level decision maker, with a high-frequency lower-level planner executing control~\cite{Dilu,RRR,DriveLikeAHuman}.  
Specifically, MPC is often employed as the lower-level motion-planning method in such architectures \cite{LanguageMPC,VLM-MPC,ASafetyPerspective}.  
However, as noted in the Introduction, these approaches do not address the scalability limitations of the motion planner itself.

\subsection{Motion Planning and Safety Assurance along with Language Models}

\looseness=-1
Language models have been applied to planning diverse, long-horizon tasks in robotics \cite{LLM4Robotics}, and many approaches---such as \cite{CodeAsPolicies,LLM2Rewards}---adopt the two-layer architecture combining an LLM with a motion planner, achieving notable success.  
Autonomous driving similarly demands long-horizon planning and adaptability to varied environments, making it a promising application domain; however, the stringent safety requirements pose significant challenges.  

Several mechanisms have been proposed to mitigate the lack of safety awareness when using an LLM as a task planner.  
Some studies passively guarantee safety by outputting control inputs that satisfy control-barrier functions~\cite{SafetyTaskPlanningRobotics,bidirectional}.  
Nonetheless, because these methods do not provide the LLM with feasibility feedback or active replanning based on the lower-level planner's results, overall safety remains dependent on the quality of the LLM's plans.  

Wang \etal combine LLM-based task planning with MPC-based motion planning.  
Similar to our approach, they feed feasibility information from the MPC back to the LLM, enabling the LLM to generate safer task plans~\cite{ASafetyPerspective}.  
However, since the LLM operates at a lower frequency than the MPC, plans deemed optimal by the LLM may be rejected due to infeasibility at the initial condition of each planning step, resulting in inefficiency.  
In contrast, our proposed method addresses this trade-off by combining passive assistance for executing tasks planned by the LVLM with active rejection feedback, thereby aiming to achieve both planning efficiency and safety.  

%% file: src/preliminary.tex
\begin{figure*}[t]
  \centering
  \includegraphics[width=\linewidth]{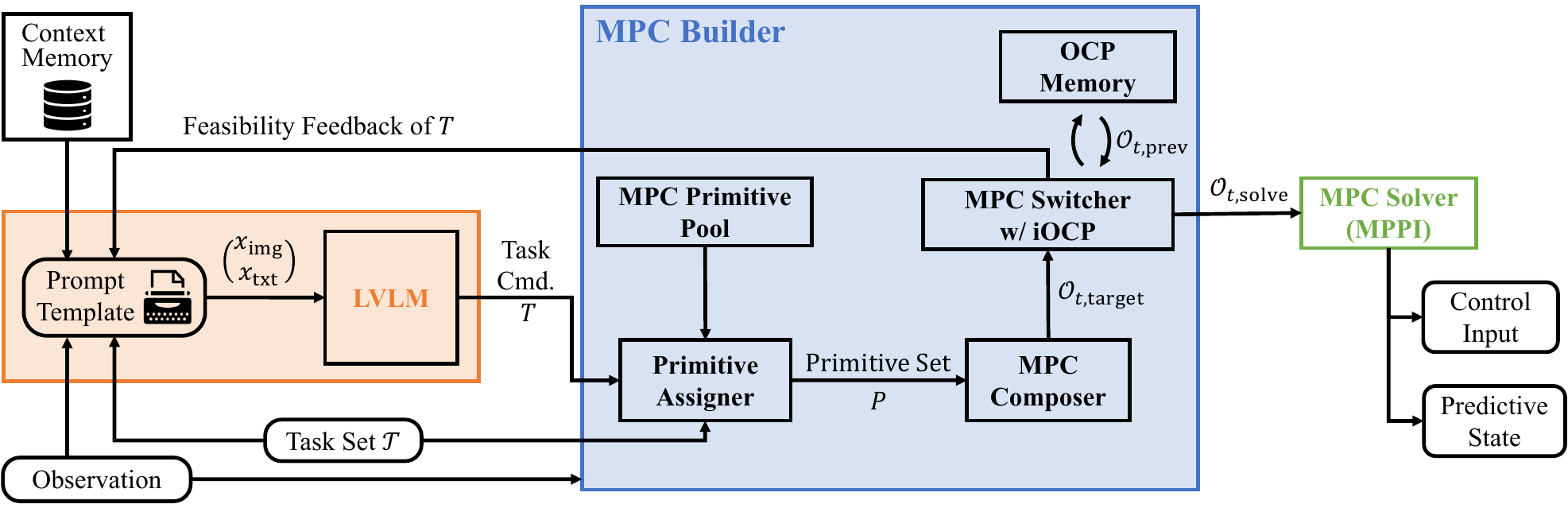}
  \caption{
  System overview of the proposed method.  
  The orange section denotes LVLM-based task planning (Section~\ref{ssec:lvlm_planning}).  
  The blue section shows OCP generation, feasibility checking, and MPC switching by MPC Builder (Section~\ref{ssec:mpc_builder_inside}).  
  MPC Builder returns feasibility feedback to the LVLM, and the OCP it produces is guaranteed feasible and is safely executed by the MPC solver to compute vehicle control inputs.  
  }
  \vspace{-1em}
  \label{fig:framework}
\end{figure*}

\section{Review of MPC Builder}
\label{sec:preliminary}

This study integrates our previously proposed MPC Builder~\cite{MPCBuilder} with the LVLM-based task planner.  
First, we provide a brief overview of MPC Builder, which is a method for generating Optimal Control Problems (OCPs) for a given task.  

\looseness=-1
The objective of MPC Builder is to execute various tasks by automatically generating the corresponding OCP $\myset{O}_t$ at time $t$ defined as:  
\begin{align}
  & \textbf{Find: } \hat{\myvec{x}}(k|t) \in \myset{X} \subset \mathbb{R}^n,\;\; \forall k\in \{1, \dots, N\},\label{eqn:mpc_x}\\
  & \qquad\quad \hat{\myvec{u}}(k|t) \in \myset{U} \subset \mathbb{R}^m,\;\; \forall k\in \{0, \dots, N-1\},\label{eqn:mpc_u}\\
  & \textbf{Min.: } J_{0:N}= \sum_{k=0}^{N} J_k\bigl(\hat{\myvec{x}}(k|t), \hat{\myvec{u}}(k|t)\bigr),\label{eqn:mpc_j}
\end{align}
\begin{align}
  & \textbf{S.t.: } \hat{\myvec{x}}(0|t) = \myvec{x}(t),\\
  & \qquad \hat{\myvec{x}}(k+1|t) = \hat{\myvec{x}}(k|t) + \myvec{f}\bigl(\hat{\myvec{x}}(k|t), \hat{\myvec{u}}(k|t)\bigr)\Delta \tau,\label{eqn:mpc_f}\\
  & \qquad \myvec{g}_{0:N}\bigl(\hat{\myvec{x}}(k|t), \hat{\myvec{u}}(k|t)\bigr) \le \myvec{0},\;\; \myvec{h}_{0:N}\bigl(\hat{\myvec{x}}(k|t), \hat{\myvec{u}}(k|t)\bigr) = \myvec{0},\label{eqn:mpc_h}
\end{align}
where the predicted values $\hat{\myvec{x}}$ and $\hat{\myvec{u}}$ are the predicted state and control-input vectors.  
$\myset{X}$ and $\myset{U}$ denote the state and control spaces, respectively.  
$N$ and $\Delta \tau$ are the length and time intervals of the prediction horizon, respectively.  
$J_{0:N}$ is the cost function that is minimized.  
$\myvec{f}$ is a state-prediction function that represents the dynamics of the control target.  
$\myvec{g}_{0:N}$ and $\myvec{h}_{0:N}$ are vectors of inequality and equality constraints, respectively.  
$n_g$ and $n_h$ are the numbers of inequality and equality constraints, respectively.  

The idea of MPC Builder is to combine \emph{MPC primitives}, which are the basic components of the OCP.  
The definition of MPC primitives is as follows:  
\begin{align}
  \myset{P} = \bigl(\myset{X}, \myvec{f}, J_{0:N}, \myvec{g}_{0:N}, \myvec{h}_{0:N}\bigr),\label{eqn:mpc_primitive_definition}
\end{align}
where it is assumed that all MPC primitives share a common input space $\myset{U}$.  
The MPC primitives are designed to be reusable and represent subtasks or control requirements, such as vehicle dynamics, lateral/longitudinal driving tasks, and safety requirements for surrounding agents, as listed in Table~\ref{tbl:primitive_formula}.  

MPC Builder generates the OCP $\myset{O}_t$ at time $t$ by combining the MPC primitives in $\myset{P}$ as follows:  
\begin{align}
  \myset{O}_t
  &= \bigl(U, \sum_{i=1}^{M} \myset{P}_i\bigr)
   = \bigl(U, ((\myset{P}_1 \oplus \myset{P}_2)\oplus\cdots)\oplus\myset{P}_M\bigr),\label{eqn:composition_primitives}
\end{align}
where $\myset{P}_i$ is the $i$-th MPC primitive, and $M$ is the number of MPC primitives.  
The operator $\oplus$ combines two MPC primitives as  
\begin{align}
  & \myset{P}_i \oplus \myset{P}_j
    = \bigl(\myset{X}, \myvec{f}, J_{0:N}, \myvec{g}_{0:N}, \myvec{h}_{0:N}\bigr),\\
  & \myset{X} = \myset{X}_i \times \myset{X}_j,\;
    \myvec{f} = \myvec{f}_i \times \myvec{f}_j,\;
    J_{0:N} = J_{0:N}^i + J_{0:N}^j,\nonumber\\
  & \myvec{g}_{0:N} = \myvec{g}_{0:N}^i \times \myvec{g}_{0:N}^j,\;
    \myvec{h}_{0:N} = \myvec{h}_{0:N}^i \times \myvec{h}_{0:N}^j.\nonumber
\end{align}

The strength of MPC Builder lies in its capability to represent various time-varying driving tasks in real time by composing a small number of MPC primitives.  
As a result, compared with designing individual MPCs for every possible driving task and environmental configuration, MPC Builder generates MPCs with fewer design elements and reduces engineering effort.  
However, the previous study~\cite{MPCBuilder} selected the required MPC primitives at each time step using a rule-based approach, which did not consider the context of the environment or the driving task.

%% file: src/method.tex
\section{LVLM Meets MPC Builder} \label{sec:method}
In this paper, we integrate the LVLM-based task planner with MPC Builder to achieve various and flexible driving tasks while considering MPC-based costs, safety, and dynamic feasibility.  

\subsection{System Overview} \label{ssec:overall_system}
Figure~\ref{fig:framework} shows the system overview of the proposed framework, which is a hierarchical structure consisting of two layers: the LVLM-based task planner and MPC Builder.  
The LVLM plans the target driving task based on the observed environmental information and passes the task command $\myset{T}$ to MPC Builder.  
The task command $\myset{T}$ is assumed to be a symbolic representation of the driving task, such as in \eqref{eq:task_command}.  
Based on the task command and the observation, MPC Builder generates a feasible OCP $\myset{O}_\mathrm{solve}$, which is solved by an optimization solver in real time.  

\looseness=-1
The LVLM task planner and MPC Builder can run separately. This works because the MPC Builder can keep doing its current task, even when it doesn't get new instructions from the LVLM. 
In contrast, the LVLM operates with low frequency, and it is difficult to issue task commands within the vehicle's control cycle.  

\subsection{LVLM-based Task Planning with Feasibility Feedback} \label{ssec:lvlm_planning}
\begin{figure}[t]
  \centering
  \includegraphics[width=\linewidth]{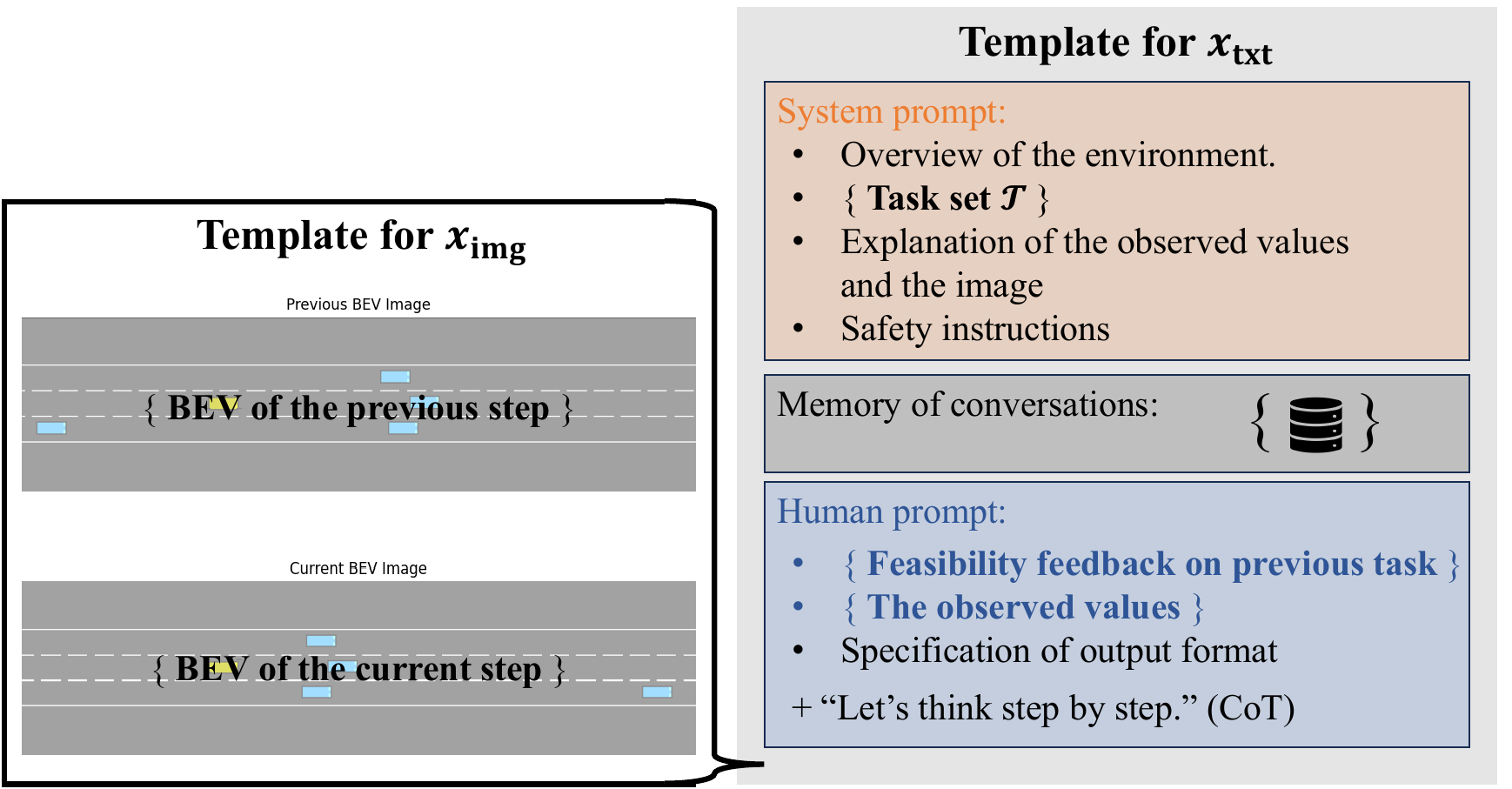}
  \caption{Template for generating text-and-image prompts to the LVLM.}
  \label{fig:prompter_template}
\end{figure}

The LVLM reasons the task command from the bird's-eye-view~(BEV) image and text prompt $(x_\mathrm{img}, x_\mathrm{txt})$ to reflect the driving situation and user intent.  
The text prompt includes system and heuristic human instructions, such as sensor observation, the format of the task command, and safety instructions to consider the surrounding traffic, as shown in Fig.~\ref{fig:prompter_template}.  
We also use prompt techniques to improve LVLM performance, specifically zero-shot Chain-of-Thought~(CoT) prompting~\cite{zeroCoT} and in-context learning~\cite{FewshotLearner} with context memory to store past reasoning and task-planning results.  

Although the LVLM can generate a task command based on the current context, the command may not always be feasible in the physical environment, \ie, the generated OCP may be infeasible.  
To reduce such flaws, we provide a binary feasibility flag to the LVLM from the MPC Builder.  
The feasibility check in MPC Builder is described in Section~\ref{ssec:mpc_builder_inside}.  

\subsection{Smooth Task Switching using Intermediate OCP} \label{ssec:iOCP}
\input{src/algorithm/intermediate_MPC}

As described in Section~\ref{ssec:lvlm_planning}, the LVLM-based task planner evaluates feasibility via feedback from MPC Builder; however, it alone cannot guarantee that the generated task command will be feasible.  
If MPC Builder simply rejects an infeasible command, the rejection may hinder the LVLM's long-term reasoning and potentially lead to a deadlock.  

\begin{figure}[t]
  \centering
  \includegraphics[width=\linewidth]{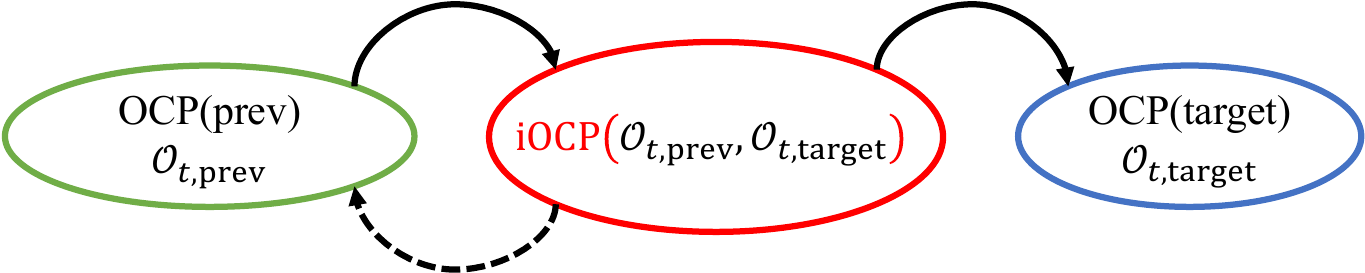}
  \caption{Transition from $\mathrm{OCP(prev)}$ to $\mathrm{OCP(target)}$ via the iOCP.  
  If $\myset{O}_{t,\mathrm{target}}$ is absolutely infeasible, the transition is abandoned, and the iOCP reverts to the safe $\myset{O}_{t,\mathrm{prev}}$.}
  \label{fig:intermediate_ocp}
  \vspace{-1em}
\end{figure}

To address this issue, we introduce the \textit{intermediate OCP} (iOCP)~\cite{IntermediateMPC} to facilitate smooth task transitions within MPC Builder.  
When the OCP changes owing to a new task command or traffic condition, MPC Builder generates a new iOCP to assist the transition from the previous OCP to the new one, as shown in Fig.~\ref{fig:intermediate_ocp}.  
The specific process is described in Algorithm~\ref{alg:intermediate_ocp}.  
In this algorithm, the iOCP is defined from the previous OCP $\myset{O}_{t,\mathrm{prev}}$ and the new OCP $\myset{O}_{t,\mathrm{target}}$ as $\mathrm{iOCP}(\myset{O}_{t,\mathrm{prev}}, \myset{O}_{t,\mathrm{target}})$.  
Here, the iOCP has the Cartesian product of the state spaces and dynamics of $\myset{O}_{t,\mathrm{prev}}$ and $\myset{O}_{t,\mathrm{target}}$, and the constraints of $\myset{O}_{t,\mathrm{prev}}$ are retained.  
The cost function of the iOCP is defined as the sum of the cost functions of $\myset{O}_{t,\mathrm{prev}}$ and $\myset{O}_{t,\mathrm{target}}$, plus a penalty function for the constraints of $\myset{O}_{t,\mathrm{target}}$.  

Consequently, the iOCP guides the ego vehicle's state toward the feasible region of the target task command while ensuring the feasibility of the previous OCP.  
Note that the iOCP requires no additional design elements other than the penalty coefficients \(\rho_\myvec{g}\) and \(\rho_\myvec{h}\).  

\subsection{OCP Generation, Feasibility Check, and MPC Switching in MPC Builder} \label{ssec:mpc_builder_inside}
We finally highlight the pipeline of MPC Builder, which generates the OCP based on the task command and observation.  
As shown in Fig.~\ref{fig:framework}, a module called \emph{Primitive Assigner} selects the necessary MPC primitive set $P=\{\myset{P}_1,\dots,\myset{P}_M\}$ from the \emph{MPC Primitive Pool} according to the task command and surrounding traffic.  
In the highway-driving experiment in this paper, we use a simple rule-based approach for assigning MPC primitives, as described in Section~\ref{ssec:implimentation} and listed in Table~\ref{tbl:primitive_formula}.  
The \emph{MPC Composer} generates a target OCP $\myset{O}_{t,\mathrm{target}}$ from the selected primitive set $P$ at time $t$ based on \eqref{eqn:composition_primitives}.  

As mentioned in Section~\ref{ssec:lvlm_planning}, MPC Builder checks the feasibility of $\myset{O}_{t,\mathrm{target}}$ and provides feedback to the LVLM.  
The \emph{MPC Switcher} verifies feasibility using the control-input sequence optimized in the previous step, $\{\hat{\myvec{u}}(k|t-1)\mid k\in\{1,\dots,N-1\}\}$, as follows:  
\begin{align}
  \begin{aligned}
    &\myset{O}_{t,\mathrm{task}}\ \text{is feasible} \;\Leftarrow \\
    &\forall k\in\{0,N-2\}\quad\Bigl(\hat{\myvec{x}}(0|t)=\myvec{x}(t)\ \wedge \\
    &\quad\hat{\myvec{x}}(k\!+\!1|t)=\hat{\myvec{x}}(k|t)+\myvec{f}(\hat{\myvec{x}}(k|t),\hat{\myvec{u}}(k\!+\!1|t-1))\Delta\tau\ \wedge \\
    &\quad\myvec{g}_k(\hat{\myvec{x}}(k|t),\hat{\myvec{u}}(k\!+\!1|t-1))\le\myvec{0}\ \wedge \\
    &\quad\myvec{h}_k(\hat{\myvec{x}}(k|t),\hat{\myvec{u}}(k\!+\!1|t-1))=\myvec{0}\Bigr).
  \end{aligned}
\end{align}
Although this feasibility check is based on the previous predictive states and is not a strict condition, it is a practical way to assess feasibility and has not caused any issues in our experiments. 

The MPC Switcher then generates an iOCP if $\myset{O}_{t,\mathrm{target}}$ is infeasible, assisting the transition as described in Section~\ref{ssec:iOCP}.  
The OCP to be solved, denoted as $O_{t,\mathrm{solve}}$, is output by the MPC Switcher, which functions according to Algorithm \ref{alg:mpc_switcher}.
We set an upper limit $n_\mathrm{max}$ on consecutive iOCP selections to prevent the system from being trapped in an iOCP.  
If $n_\mathrm{iOCP}>n_\mathrm{max}$, MPC Builder rejects the task switch, reverts to $\myset{O}_\mathrm{prev}$, and marks the task command infeasible, as indicated in Fig.~\ref{fig:intermediate_ocp}.  
Because the constraints of the iOCP and previous OCP are the same, this reversion is guaranteed to be feasible.  
This decision is indicated as $\mathrm{is\_rejected}$ in Algorithm 2, and is also fed back to the LVLM as a rejection.  

\input{src/algorithm/mpc_switcher}

As a result, MPC Builder generates a feasible OCP $\myset{O}_{t,\mathrm{solve}}$ even when the task command is infeasible.  
Moreover, we achieve smooth and continuous task switching, although the LVLM's task planning is coarse and asynchronous.

%% file: src/algorithm/intermediate_MPC.tex
\begin{algorithm}[t]
\small
  \caption{Intermediate OCP between two OCPs}
  \label{alg:intermediate_ocp}
  \begin{algorithmic}[1]
  \Require{Coefficients of the penalty functions, $\rho_\myvec{g}$ and $\rho_\myvec{h}$}
  \Function{iOCP}{$\myset{O}_A$, $\myset{O}_B$}
        \State{}\Comment{$\myset{O}_A = \myset{O}_{t, \mathrm{prev}}, \myset{O}_B = \myset{O}_{t, \mathrm{target}}$ in this paper}
        \State{$( \myset{U}, \myset{X}_A, \myvec{f}_A, J_{0:N}^A, \myvec{g}_{0:N}^A, \myvec{h}_{0:N}^A ) \gets \myset{O}_A$}
        \State{$( \myset{U}, \myset{X}_B, \myvec{f}_B, J_{0:N}^B, \myvec{g}_{0:N}^B, \myvec{h}_{0:N}^B ) \gets \myset{O}_B$}
        \State{$\myset{X}_\mathrm{AB} \gets \myset{X}_A \times \myset{X}_B$}\label{step:state_space}
        \Comment{cartesian product of state space}
        \State{$\myvec{x}_{\mathrm{AB}} \gets \left( \myvec{x}_A, \myvec{x}_B \right) \in \myset{X}_{\mathrm{AB}};\ \myvec{u} \in \myset{U}$}
        % \Comment{get ordered state vector}
        \State{$\myvec{f}_{\mathrm{AB}}(\myvec{x}_{\mathrm{AB}}, \myvec{u}) = \left(\myvec{f}_{A}(\myvec{x}_{A}, \myvec{u}), \myvec{f}_{B}(\myvec{x}_{B}, \myvec{u}) \right)$}\label{step:prediction_model}\\
        \Comment{extend state prediction function}
        \State{$P_k(\myvec{x}_k^B, u) = \rho_\myvec{g} \max\{0, \myvec{g}_{k}^{B}(\myvec{x}_k^B, u)\}^2 + \rho_\myvec{h}\left\lVert\myvec{h}_k^B(\myvec{x}_k^B, u)\right\rVert^2$}\\
        \Comment{define penalty function from constrains of $\myset{O}_B$}
        \State{$ J_{0:N}^{\mathrm{AB}}(\myvec{x}_{\mathrm{AB}}, \myvec{u}) = J_{0:N}^{A}(\myvec{x}_{A}, \myvec{u}) + \sum_{k=0}^{N-1} P_k(\myvec{x}_B^k, u)$}\label{step:cost_func}\\
        \Comment{add penalty to cost function of $\myset{O}_A$}
        \State{$\myvec{g}_{0:N}^{\mathrm{AB}}(\myvec{x}_{\mathrm{AB}}, \myvec{u}) = \myvec{g}_{0:N}^{A}(\myvec{x}_{A}, \myvec{u}) $}\label{step:ineq_constraint}\\
        \Comment{same inequality constrants as $\myset{O}_B$}
        \State{$\myvec{h}_{0:N}^{\mathrm{AB}}(\myvec{x}_{\mathrm{AB}}, \myvec{u}) = \myvec{h}_{0:N}^{A}(\myvec{x}_{A}, \myvec{u})$}\label{step:eq_contraint}\\
        \Comment{same equality constrants as $\myset{O}_A$}
        \State{$\myset{O}_{AB} \gets ( \myset{U}, \myset{X}_{AB},  \myvec{f}_{AB}, J_{AB}, \myvec{g}_{AB}, \myvec{h}_{AB} )$}\\
  \Return{$\myset{O}_{AB}$}
  \EndFunction
  \end{algorithmic}
\end{algorithm}

%% file: src/algorithm/mpc_switcher.tex
\begin{algorithm}[t]
\small
  \caption{MPC Switcher}
  \label{alg:mpc_switcher}
  \begin{algorithmic}[1]
  \Require{The OCP from the previous time step, $\myset{O}_{t,\mathrm{prev}}$, the consecutive execution count of iOCP, $n_\mathrm{iOCP}$, and the upper limit of $n_\mathrm{iOCP}$, $n_\mathrm{max}$}
  \Function{mpc\_switcher}{$\myset{O}_{t,\mathrm{task}}$}
        \If{$\myset{O}_{t,\mathrm{task}}$ is feasible}
            \State{$\myset{O}_{t,\mathrm{solve}} \gets \myset{O}_{t,\mathrm{task}}$}
            \State{$\myset{O}_{t,\mathrm{prev}} \gets \myset{O}_{t,\mathrm{task}}$}
            \State{$n_\mathrm{iOCP} \gets 0$}
            \State{$\mathrm{is\_rejected} \gets \mathrm{False}$ }
        \Else
            \If{$n_\mathrm{iOCP} < n_\mathrm{max}$}
                \State{$\myset{O}_{t,\mathrm{solve}} \gets \mathrm{iOCP}(\myset{O}_{t, \mathrm{prev}},\myset{O}_{t,\mathrm{task}})$}
                \State{$n_\mathrm{iOCP} \gets n_\mathrm{iOCP}+1$}
                \State{$\mathrm{is\_rejected} \gets \mathrm{False}$ }
            \Else
                \State{$\myset{O}_{t,\mathrm{solve}} \gets \myset{O}_{t, \mathrm{prev}}$}
                \State{$n_\mathrm{iOCP} \gets 0$}
                \State{$\mathrm{is\_rejected} \gets \mathrm{True}$ }
            \EndIf
        \EndIf\\
  \Return{$\myset{O}_{t,\mathrm{solve}}, \mathrm{is\_rejected}$}
  \EndFunction
  \end{algorithmic}
\end{algorithm}

%% file: src/experiment.tex
\section{Experiments} \label{sec:experiment}

To demonstrate the safety-assurance mechanism of the proposed framework, we conducted numerical experiments in a simulated driving environment, where the ego vehicle drives in a congested three-lane highway scenario based on the HighwayEnv simulator~\cite{HighwayEnv}.

\subsection{Experimental Setup} \label{ssec:implimentation}

We use the GPT-4o model~\cite{gpt-4o} as the LVLM. Since we access the model through the OpenAI API and the delay time is not reproducible, we implemented a synchronous process to pause the simulation until the LVLM produced a response.

In the highway-driving scenario, we feed the following instruction in the user-instruction part of the text prompt (Fig.~\ref{fig:prompter_template}): \ie, ``\emph{You should now act as a mature driving assistant who can give accurate and correct advice for a human driver in complex highway driving scenarios. The passengers say, "I'm in a hurry."}''  
This prompt encourages the LVLM to weave through the congested road by lane changing.

As described in Section~\ref{ssec:lvlm_planning}, the LVLM generates a task command $T$.
Specifically, the LVLM outputs a symbolic command $T$ that indicates the desired driving task as
\begin{align}
    \myset{T} = \{\texttt{LANE\_LEFT}, \texttt{IDLE}, \texttt{LANE\_RIGHT}\}.\label{eq:task_command}
\end{align}

\looseness=-1
As the MPC Builder, we implemented six MPC primitives to execute the task commands and ensure safety for surrounding vehicles, as shown in Table~\ref{tbl:primitive_formula}.
While our implementation focuses on the highway-driving scenario, the MPC Builder framework has been validated across multiple scenarios in \cite{MPCBuilder}, showing its potential applicability to diverse tasks and environments.

The Primitive Assigner selects a lateral task primitive based on the task command $T \in \myset{T}$ received from the LVLM.
It selects a longitudinal task primitive between Constant Speed~(CS) and Adaptive Cruise Control~(ACC) using a simple rule; CS is chosen when the leading vehicle is located far ahead of the ego vehicle, whereas ACC is chosen when the leading vehicle is as close as $2d_\mathrm{acc}\,[\mathrm{m}]$ to the ego vehicle.
For safety primitives corresponding to surrounding vehicles, all vehicles in the vicinity of the ego vehicle are selected, up to a maximum of six.

\input{src/table/formulation_mpc_primitive}

Regarding the MPC solver, to improve numerical stability, we adopted the sample-based method MPPI~\cite{MPPI} instead of the C/GMRES method used in \cite{MPCBuilder}.
Because constraints cannot be handled as hard constraints, the cost function $C_k$ provided to MPPI incorporates the constraint functions as penalty terms:
\begin{align}
    C_k(\hat{\myvec{x}}, \hat{\myvec{u}}) = J_k(\hat{\myvec{x}}, \hat{\myvec{u}}) + \mu \!\left(\mathbf{1}[\myvec{g}_k(\hat{\myvec{x}}, \hat{\myvec{u}}) > \myvec{0}] + \mathbf{1}[\myvec{h}_k(\hat{\myvec{x}}, \hat{\myvec{u}}) \neq \myvec{0}]\right),
\end{align}
where $\mathbf{1}[\cdot]$ is an indicator function that returns 1 if the given proposition is true and 0 otherwise.
The coefficient $\mu$ is set to a sufficiently large value ($\mu=100$) to treat the constraints as approximately hard constraints in this experiment.
The prediction horizon is configured with $N = 20$ steps.
The time step is set as $\Delta \tau = 0.05\,\mathrm{s}$.
The control input is defined as $u = (a, \delta) \in \myset{U}$, where $a$ represents acceleration and $\delta$ the steering angle.
The variances for input sampling used in MPPI are set to 2.0 and 0.01 for acceleration and steering angle, respectively.

\subsection{Evaluation Setup}\label{ssec:baseline_metrics}

To evaluate the effectiveness of the proposed bidirectional communication framework between the LVLM and MPC Builder, we compare the proposed method, called \textbf{LVLM-MPCBuilder}, with two baseline methods:
\begin{itemize}
  \item \textbf{LVLM2MPC} executes the tasks planned by the LVLM directly by solving the target OCP generated by the MPC Builder without iOCP assistance. No feasibility feedback is sent to the LVLM, resulting in strictly one-way communication. This baseline is similar to \cite{LanguageMPC}.
  \item \textbf{LVLM2PID} utilizes a PID controller built into HighwayEnv to execute the tasks planned by the LVLM, which decides a task command $T$ from the command set
        \begin{align}
            \myset{T} = &\{\texttt{LANE\_LEFT}, \texttt{IDLE}, \texttt{LANE\_RIGHT}, \nonumber\\ &\texttt{FASTER}, \texttt{SLOWER}\}.
        \end{align}
        LVLM2PID also sends no feasibility feedback to the LVLM. In contrast to the proposed method and LVLM2MPC, the PID controller does not perform trajectory planning that accounts for safety constraints, similar to \cite{Dilu,DriveLikeAHuman}.
\end{itemize}
In all methods, including the proposed method, we use the same text and image prompts for the LVLM.
Note that the prompt includes a safety instruction asking the LVLM to maintain a safe distance $d_{\rm{safe}}^{\rm{acc}}$ from surrounding vehicles during lane changes.
For LVLM-MPCBuilder, every time the LVLM performed a single task-planning step, MPC Builder executed an additional 30 control steps beyond the iOCP steps, and the maximum number of iOCP steps was set to $n_\mathrm{max}=50$.
For LVLM2PID, the LVLM carried out task planning at a frequency of $1\,\mathrm{Hz}$.

We utilize the following safety metrics to evaluate each method:
\begin{itemize}
  \item \textbf{Success rate: } the proportion of episodes that terminate without collision.
  \item \textbf{Safe lane-changing rate: } the proportion of lane changes that maintain the safety constraint of a distance of at least $d_\mathrm{safe}^\mathrm{acc}$ from the preceding and following vehicles in the target lane.
\end{itemize}

\subsection{Simulation Results}
\subsubsection{Comparison with Baseline Methods}
Table~\ref{tbl:result} summarizes the quantitative evaluation results with 30 randomly generated initial positions of the ego vehicle and surrounding vehicles during the $50\,\mathrm{s}$ simulation time.

As shown in Table~\ref{tbl:result}, the baseline method LVLM2PID exhibits lower success rates than the other methods because it does not perform trajectory planning that accounts for safety constraints.
Although the prompt includes the safety instruction to maintain a safe distance $d_\mathrm{acc}$ from leading vehicles, LVLM2PID often collides with leading vehicles, resulting in a low success rate.
This indicates that the LVLM does not reliably enforce quantitative constraints, highlighting the necessity of hard-constraint enforcement by the motion planner.

\begin{figure}[t]
  \centering
  \includegraphics[width=\linewidth]{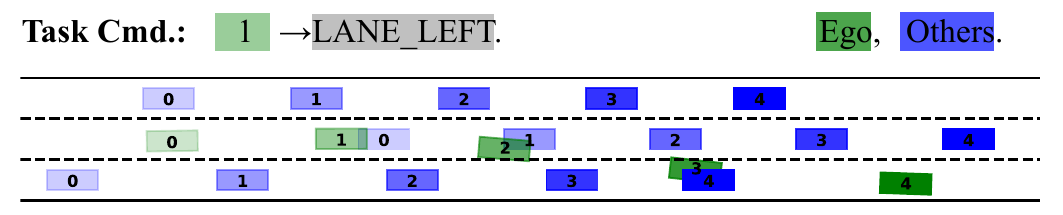}
  \caption{The LVLM's reckless lane-change decision was executed faithfully, causing the following vehicle to decelerate.}
  \label{fig:nearMiss_mpc}
\end{figure}

In contrast, the proposed LVLM-MPCBuilder demonstrates both a success rate of $100\%$ and a safe lane-changing rate of $100\%$, indicating that the proposed method successfully ensures safety throughout the simulation.
Among the 133 lane-change decisions made by the LVLM, 9 lane changes were assisted or rejected by the iOCP through feasibility checking and state guidance.
Hence, only $(133-9)/133 = 93.2\%$ of the lane changes decided by the LVLM satisfied the safety constraints at the moment of decision, indicating that the LVLM does not strictly obey the safety instructions in the prompt.
Indeed, in LVLM2MPC, which omits feasibility feedback and iOCP-based task switching, the safe lane-changing rate drops to $94.8\%$, and a dangerous, near-miss lane change caused the following vehicle to decelerate, as shown in Fig.~\ref{fig:nearMiss_mpc}.

\begin{figure}[t]
  \centering
  \includegraphics[width=\linewidth]{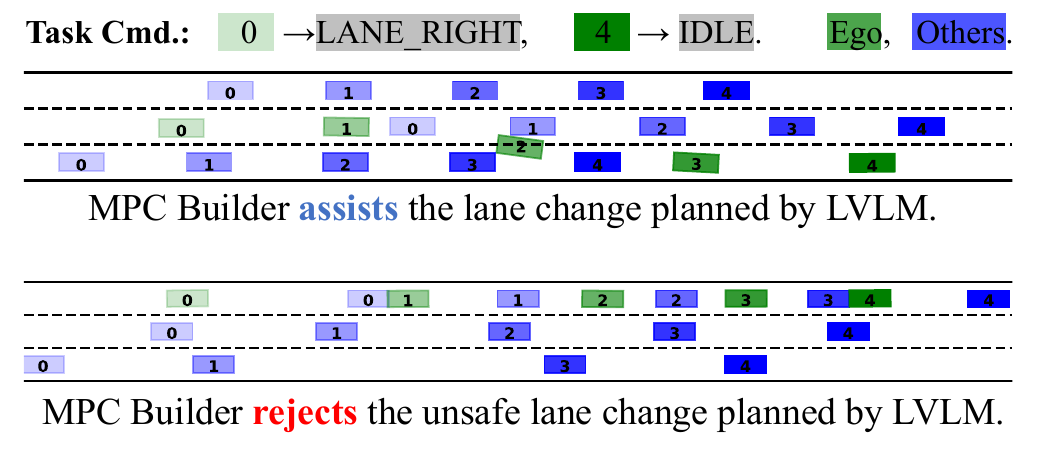}
  \caption{In LVLM-MPCBuilder, MPC Builder assists or rejects task plans generated by the LVLM that violate safety constraints, thereby ensuring safety.}
  \label{fig:exp_iOCP}
\end{figure}

Figure~\ref{fig:exp_iOCP} shows example trajectories in the simulation using LVLM-MPCBuilder.
Our method assisted LVLM-based lane changes via the iOCP, which could make the lane change feasible by adjusting the ego vehicle's speed and steering angle.
MPC Builder sometimes rejected the LVLM's decision and fed the infeasibility back to the LVLM, as described in Section~\ref{ssec:mpc_builder_inside}.

\input{src/table/experiment_result_methods}

\subsubsection{Ablation Study for the iOCP-based Task Switching}
\begin{figure}[t]
  \centering
  \includegraphics[width=0.8\linewidth]{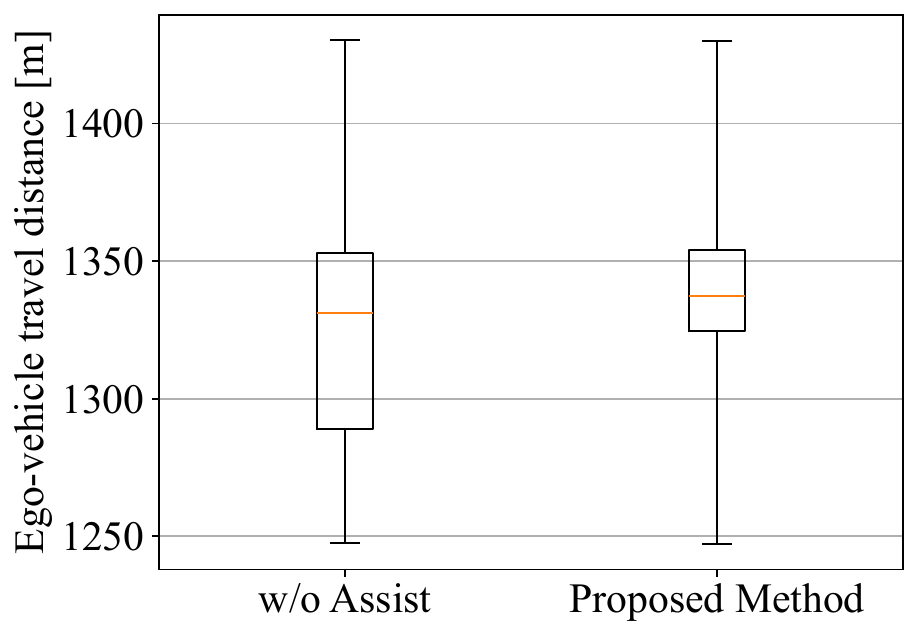}
  \caption{Ablation study for iOCP-based task-switching assistance. Comparison of ego-vehicle travel distances over all 30 episodes with and without iOCP assistance.}
  \label{fig:efficiency}
\end{figure}

We conducted an ablation study to evaluate the iOCP-based task-switching assistance mechanism in the proposed method.
Figure~\ref{fig:efficiency} compares the ego-vehicle travel distances over all 30 episodes with and without iOCP assistance.
Without iOCP assistance, episodes cover shorter distances, resulting in a lower average travel distance.
This indicates that iOCP assistance not only reduces iterative replanning by the LVLM but also mitigates deadlocks caused by rejections without alternative task plans.
Indeed, without iOCP assistance there were 17 rejections, whereas with iOCP only two tasks were rejected across all 30 episodes (Table~\ref{tbl:result}), suggesting that many reckless task plans do not require outright rejection.

\subsubsection{Ablation Study for the Safety Instruction in the Prompt}
We also conducted an ablation study to evaluate the robustness of the proposed method against prompt engineering.
In this study, we omitted the \emph{Safety Instructions} in the LVLM prompt, which specify the safe distance $d_{\rm{safe}}^{\rm{lc}}$ for lane changes described in Section~\ref{ssec:implimentation}.

\begin{figure}[t]
  \centering
  \includegraphics[width=\linewidth]{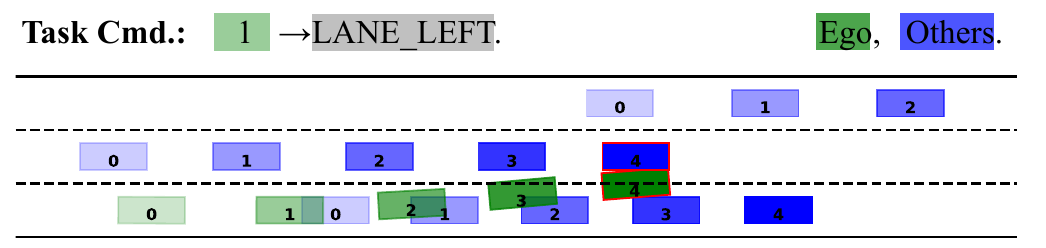}
  \caption{A collision resulting from a dangerous lane-change decision by LVLM2MPC without safety-constraint knowledge.}
  \label{fig:collision_mpc_wo_safetInstruction}
\end{figure}

Table~\ref{tbl:result_wo_safetyInstruction} summarizes the results of this ablation study.
The baseline methods, LVLM2PID and LVLM2MPC, exhibit significant decreases in success rate and safe lane-changing rate compared with the case that includes safety instructions.
An example collision from LVLM2MPC without safety instructions is shown in Fig.~\ref{fig:collision_mpc_wo_safetInstruction}.
This occurs because the LVLM's reckless lane-change decisions are executed directly by the underlying MPC or PID controller, leading to a high likelihood of collisions.
In contrast, LVLM-MPCBuilder maintains a success rate of $100\%$ and a safe lane-changing rate of $100\%$ even without the safety instruction prompt, demonstrating that the proposed method can ensure safety when the LVLM lacks explicit knowledge of safety constraints.

\input{src/table/experiment_result_instruction}

%% file: src/table/formulation_mpc_primitive.tex
\begin{table*}[t]
% \footnotesize
\centering
\caption{Formulations of MPC primitives}
% \vspace{-2mm}
\label{tbl:primitive_formula}
\begin{threeparttable}
\begingroup
\renewcommand{\arraystretch}{1.3}
\begin{tabular}{c|c|ccccc}
\hline
\rowcolor[HTML]{C0C0C0} 
  Type &
  name &
  $\myset{X}$ &
  $J_{0:N}$ &
  $\myvec{f}$ &
  $\myvec{g}_{0:N}$ &
  $\myvec{h}$ \\ \hline
\cellcolor[HTML]{FFFFC7} Ego Dynamics &
  \cellcolor[HTML]{FFFFC7}KBM &
  $X_{\rm{ego}}$ &
  0 &
  $\myvec{f}_{\rm{KBM}}$ &
  $\emptyset$ &
  $\emptyset$ \\ \hline  % \cline{2-7} 
% \multirow{-2}{*}{\cellcolor[HTML]{FFFFC7}\begin{tabular}[c]{@{}c@{}}Ego\\Dynamics\end{tabular}} &
%   \cellcolor[HTML]{FFFFC7}DBM &
%   $X_{\rm{ego}}$ &
%   0 &
%   $\myvec{f}_{\rm{DBM}}$ &
%   $\emptyset$ &
%   $\emptyset$ \\ \hline
\cellcolor[HTML]{FFCE93} &
  \cellcolor[HTML]{FFCE93}LK &
  $\emptyset$ &
  $\{\|y\|^2+\|\theta\|^2+\|\dot{\theta}\|^2+\|\delta\|^2+\|\dot{\delta}\|^2\}_{Q_{\rm{lk}}}$ &
  $\emptyset$ &
  $ [y_{\rm{min}} - y, y - y_{\rm{max}}, \delta_{\rm{min}} - \delta,  \delta - \delta_{\rm{max}} ]$ &
  $\emptyset$ \\ \cline{2-7} 
% \cellcolor[HTML]{FFCE93} &
%   \cellcolor[HTML]{FFCE93}Avoid &
%   $\emptyset$ &
%   $\{\|\theta\|^2+\|\dot{\theta}\|^2+\|\delta\|^2+\|\dot{\delta}\|^2\}_{Q_{\rm{av}}}$ &
%   $\emptyset$ &
%   \begin{tabular}[c]{@{}c@{}}$ [y_{\rm{min}} - y,  y -  y_{\rm{max}}, \delta_{\rm{min}} - \delta,  \delta - \delta_{\rm{max}},$\\ $d_{\rm{safe}}^{\rm{av}} - \sqrt{(x-x_o)^2+(y-y_o)^2} ]$\end{tabular} &
%   $\emptyset$ \\ \cline{2-7} 
\multirow{-2}{*}{\cellcolor[HTML]{FFCE93}\begin{tabular}[c]{@{}c@{}}Lateral\\Task\end{tabular}} &
  \cellcolor[HTML]{FFCE93} LC &
  $\emptyset$ &
  $\{\|y-y_{\rm{ref}}\|^2+\|\theta\|^2+\|\dot{\theta}\|^2+\|\delta\|^2+\|\dot{\delta}\|^2\}_{Q_{\rm{lc}}}$ &
  $\emptyset$ &
  \begin{tabular}[c]{@{}c@{}}$[y_{\rm{min}} - y, y - y_{\rm{max}}, \delta_{\rm{min}} - \delta, \delta - \delta_{\rm{max}},$\\ $d_{\rm{safe}}^{\rm{lc}} - \|x-x_{\rm{pv}}\|]$\end{tabular} &
  $\emptyset$ \\ \hline
\cellcolor[HTML]{96FFFB} &
  \cellcolor[HTML]{96FFFB} CS &
  $\emptyset$ &
  $\{\|v-v_{\rm{ref}}\|^2+\|a\|^2+\|\dot{a}\|^2\}_{Q_{\rm{cs}}}$ &
  $\emptyset$ &
  $[ a_{\rm{min}} - a, a - a_{\rm{max}} ]$ &
  $\emptyset$ \\ \cline{2-7} 
\multirow{-2}{*}{\cellcolor[HTML]{96FFFB}\begin{tabular}[c]{@{}c@{}}Longitudinal\\Task\end{tabular}} &
  \cellcolor[HTML]{96FFFB}ACC &
  $\emptyset$ &
  $\{\|x-x_{\rm{pv}}-d_{\rm{acc}}\|^2+\|a\|^2+\|\dot{a}\|^2\}_{Q_{\rm{acc}}}$ &
  $\emptyset$ &
  $[a_{\rm{min}} - a, a - a_{\rm{max}}, d_{\rm{safe}}^{\rm{acc}} - \|x-x_{\rm{pv}}\|]$ &
  $\emptyset$ \\ \hline  % \cline{2-7} 
% \multirow{-3}{*}{\cellcolor[HTML]{96FFFB}\begin{tabular}[c]{@{}c@{}}Longitudinal\\Task\end{tabular}} &
%   \cellcolor[HTML]{96FFFB}Stop &
%   $\emptyset$ &
%   $\{\|x-x_{\rm{stop}}\|^2+\|v\|^2+\|a\|^2+\|\dot{a}\|^2\}_{Q_{\rm{stop}}}$ &
%   $\emptyset$ &
%   $[ a_{\rm{min}} - a, a - a_{\rm{max}}, d_{\rm{safe}}^{\rm{stop}} - \|x-x_{\rm{stop}}\| ]$ &
%   $\emptyset$ \\ \hline
% \cellcolor[HTML]{9AFF99} &
%   \cellcolor[HTML]{9AFF99}PED &
%   $X_{\rm{ped}}$ &
%   $\{\|v\|^2\}_{Q_{\rm{ped}}}$ &
%   $\myvec{f}_{\rm{ped}}$  &
%   $[ d_{\rm{safe}}^{\rm{ped}} - \sqrt{(x-x_{\rm{ped}})^2+(y-y_{\rm{ped}})^2} ]$ &
%   $\emptyset$ \\ \cline{2-7} 
\cellcolor[HTML]{9AFF99} Safety &
  \cellcolor[HTML]{9AFF99}PV &
  $X_{\rm{pv}}$ &
  0 &
  $\myvec{f}_{\rm{pv}}$  &
  $[ d_{\rm{safe}}^{\rm{pv}} - \sqrt{(x-x_{\rm{pv}})^2+(y-y_{\rm{pv}})^2} ]$ &
  $\emptyset$ \\ \hline  % \cline{2-7} 
% \multirow{-3}{*}{\cellcolor[HTML]{9AFF99}\begin{tabular}[c]{@{}c@{}}Safety\end{tabular}} &
%   \cellcolor[HTML]{9AFF99}CV &
%   $X_{\rm{cv}}$ &
%   $\{\|v\|^2\}_{Q_{\rm{cv}}}$ &
%   $\myvec{f}_{\rm{cv}}$  &
%   $[ d_{\rm{safe}}^{\rm{cv}} - \sqrt{(x-x_{\rm{cv}})^2+(y-y_{\rm{cv}})^2} ]$ &
%   $\emptyset$ \\ \hline
\end{tabular}
\endgroup
\begin{tablenotes}[flushleft]
\item[1] KBM means kinematic bicycle models. Lane keep (LK) and lane change (LC) are in lateral task primitive. Constant speed (CS) and adaptive-cruise-control (ACC) are in longitudinal task primitive. Parallel vehicle (PV) is safety primitive.
\item[2] $\myset{X}_{\rm{ego}}$ is a state space of ego vehicle; $\myset{X}_{\rm{ego}} = [x, y, \theta, v]$. Here, ($x$, $y$, $\theta$) is an ego vehicle pose. 
$v$, $a$, and $\delta$ are the ego vehicle's speed, acceleration, and tire steer angle. 
$\myset{X}_{\rm{pv}} = [x_{\rm{pv}}, y_{\rm{pv}}, v_x^{\rm{pv}}, v_y^{\rm{pv}}]$ is the state space of a parallel running vehicle where $(x_{\rm{pv}}, y_{\rm{pv}})$ and $(v_x^{\rm{pv}}, v_y^{\rm{pv}})$ are position and velocity.
$\myvec{f}_{\rm{pv}}$ is the constant velocity model.
$v_{\rm{ref}}$ and $y_{\rm{ref}}$ are given the reference speed and the y-coordinate value of the lane center.
$d_{\rm{acc}}$ is a target relative distance in ACC.
$Q_*$ represents the given weight coefficient vectors. $(*_{\rm{min}}, *_{\rm{max}})$ and $d_{\rm{safe}^*}$ are given min./max. values and safety distances.
\end{tablenotes}
\end{threeparttable}
% \vspace{-3mm}
\end{table*}

%% file: src/table/experiment_result_methods.tex
\begin{table}[t]
\small
\centering
\caption{Comparison of experimental results between the baseline methods and the proposed method.}
\begin{tabular}{c|c}
\hline\hline
\multicolumn{2}{l}{LVLM2PID} \\ \hline
No. of LVLM task planning steps & 728 \\ 
No. of lane change decisions & 138 \\
Success rate   & 4/30 \\
Safe lane-changing rate [\%] & 91.3 \\

\hline\hline
\multicolumn{2}{l}{LVLM2MPC} \\ \hline
No. of LVLM task planning steps & 860 \\ 
No. of lane change decisions & 134 \\
Success rate   & 30/30 \\
Safe lane-changing rate [\%] & 94.8 \\

\hline\hline
\multicolumn{2}{l}{LVLM-MPCBuilder (Proposed Method)} \\ \hline
No. of LVLM task planning steps & 832 \\ 
No. of lane change decisions & 133 \\
No. of lane changes assisted by MPC Builder & 7 \\
No. of lane changes rejected by MPC Builder & 2 \\
Success rate   & \textbf{30/30} \\
Safe lane-changing rate [\%] & \textbf{100} \\
\hline  
\end{tabular}
\label{tbl:result}
\end{table}

%% file: src/table/experiment_result_instruction.tex
\begin{table}[t]
\small
\centering
\caption{Ablation Study: Without Safety Instructions in the Prompt.}
\begin{tabular}{c|c}
\hline\hline
\multicolumn{2}{l}{LVLM2PID} \\ \hline
% No. of LVLM task planning steps & 667 \\ 
% No. of lane change decisions & 131 \\
$\Delta$success rate &  -3/30 \\
% $\Delta$ safe lane-changing rate [\%] & 76.3 \\ 
$\Delta$safe lane-changing rate [\%] & -15.0 \\ 

\hline\hline
\multicolumn{2}{l}{LVLM2MPC} \\ \hline
% No. of LVLM task planning steps & 727 \\ 
% No. of lane change decisions & 144 \\
$\Delta$success rate & -2/30 \\
% $\Delta$ safe lane-changing rate [\%] & 59.7 \\
% $\Delta$ safe lane-changing rate [\%] & -37.1 \\
$\Delta$safe lane-changing rate [\%] & -20.8 \\

\hline\hline
\multicolumn{2}{l}{LVLM-MPCBuilder (Proposed Method)} \\ \hline
% No. of LVLM task planning steps & 782 \\ 
No. of lane change decisions & 158 \\
No. of lane changes assisted by MPC Builder & 52 \\
No. of lane changes rejected by MPC Builder & 10 \\
$\Delta$success rate & \textbf{0} \\
$\Delta$safe lane-changing rate [\%] & \textbf{0} \\
\hline
\end{tabular}
\label{tbl:result_wo_safetyInstruction}
\end{table}

%% file: src/conclusion.tex
\section{Conclusion} \label{sec:conclusion}

This paper proposes an AD planning framework that combines LVLM-based task planning with MPC Builder-based motion planning.
Our approach leverages the strengths of both components: the LVLM's ability to generate high-level task plans and the MPC Builder's capability to produce safe and task-scalable motion plans.
Simulation experiments demonstrated that the proposed method significantly improves safety compared with baseline methods.  
Because MPC Builder offers high scalability with respect to driving tasks and scenarios, the proposed system has the potential to exploit the LVLM's versatility without loss of generality.  

Our proposed method has several limitations. A primary limitation is the latency of the LVLM. This not only poses a challenge to real-time processing but may also impede immediate responses to abrupt environmental changes.
Another limitation is the lack of a systematic method for tuning the cost weights and penalty factors in the MPC. Although this is an inherent challenge in OCPs themselves, the proper tuning of these parameters becomes particularly difficult in complex tasks and scenarios.
As future work, we will validate the framework under diverse and complex scenarios and conduct real-world, real-time experiments to clarify and resolve the challenges posed by these limitations.

%% file: main.bbl
\begin{thebibliography}{10}
\providecommand{\url}[1]{#1}
\csname url@samestyle\endcsname
\providecommand{\newblock}{\relax}
\providecommand{\bibinfo}[2]{#2}
\providecommand{\BIBentrySTDinterwordspacing}{\spaceskip=0pt\relax}
\providecommand{\BIBentryALTinterwordstretchfactor}{4}
\providecommand{\BIBentryALTinterwordspacing}{\spaceskip=\fontdimen2\font plus
\BIBentryALTinterwordstretchfactor\fontdimen3\font minus \fontdimen4\font\relax}
\providecommand{\BIBforeignlanguage}[2]{{%
\expandafter\ifx\csname l@#1\endcsname\relax
\typeout{** WARNING: IEEEtran.bst: No hyphenation pattern has been}%
\typeout{** loaded for the language `#1'. Using the pattern for}%
\typeout{** the default language instead.}%
\else
\language=\csname l@#1\endcsname
\fi
#2}}
\providecommand{\BIBdecl}{\relax}
\BIBdecl

\bibitem{gpt-4o}
R.~Islam and O.~M. Moushi, ``{GPT-4o}: The cutting-edge advancement in multimodal llm,'' \emph{Authorea Preprints}, 2024.

\bibitem{MPC_switch}
Q.~Wang, B.~Ayalew, and T.~Weiskircher, ``Predictive maneuver planning for an autonomous vehicle in public highway traffic,'' \emph{IEEE Transactions on Intelligent Transportation Systems}, vol.~20, no.~4, pp. 1303--1315, 2018.

\bibitem{MPC_MixedInteger}
C.~Liu, S.~Lee, S.~Varnhagen, and H.~E. Tseng, ``Path planning for autonomous vehicles using model predictive control,'' in \emph{Intelligent Vehicles Symposium}.\hskip 1em plus 0.5em minus 0.4em\relax IEEE, 2017, pp. 174--179.

\bibitem{LanguageMPC}
\BIBentryALTinterwordspacing
H.~Sha, Y.~Mu, Y.~Jiang, L.~Chen, C.~Xu, P.~Luo, S.~E. Li, M.~Tomizuka, W.~Zhan, and M.~Ding, ``Languagempc: Large language models as decision makers for autonomous driving,'' 2023. [Online]. Available: \url{https://arxiv.org/abs/2310.03026}
\BIBentrySTDinterwordspacing

\bibitem{bidirectional}
Z.~Ma, Q.~Sun, and T.~Matsumaru, ``Bidirectional planning for autonomous driving framework with large language model,'' \emph{Sensors}, vol.~24, no.~20, p. 6723, 2024.

\bibitem{ASafetyPerspective}
Y.~Wang, R.~Jiao, S.~S. Zhan, C.~Lang, C.~Huang, Z.~Wang, Z.~Yang, and Q.~Zhu, ``Empowering autonomous driving with large language models: A safety perspective,'' in \emph{ICLR 2024 Workshop on Large Language Model Agents}, 2024.

\bibitem{MPCBuilder}
K.~Honda, H.~Okuda, T.~Suzuki, and A.~Ito, ``Mpc builder for autonomous drive: Automatic generation of mpcs for motion planning and control,'' in \emph{Intelligent Vehicles Symposium}.\hskip 1em plus 0.5em minus 0.4em\relax IEEE, 2023, pp. 1--8.

\bibitem{long-tailed}
A.~Jain, L.~Del~Pero, H.~Grimmett, and P.~Ondruska, ``Autonomy 2.0: Why is self-driving always 5 years away?'' \emph{arXiv preprint arXiv:2107.08142}, 2021.

\bibitem{RuleLaneChange}
C.~F. Choudhury and M.~E. Ben-Akiva, ``Modelling driving decisions: a latent plan approach,'' \emph{Transportmetrica A: Transport Science}, vol.~9, no.~6, pp. 546--566, 2013.

\bibitem{DLLaneChange}
D.-F. Xie, Z.-Z. Fang, B.~Jia, and Z.~He, ``A data-driven lane-changing model based on deep learning,'' \emph{Transportation research part C: emerging technologies}, vol. 106, pp. 41--60, 2019.

\bibitem{RLLaneChange}
F.~Ye, X.~Cheng, P.~Wang, C.-Y. Chan, and J.~Zhang, ``Automated lane change strategy using proximal policy optimization-based deep reinforcement learning,'' in \emph{Intelligent Vehicles Symposium}.\hskip 1em plus 0.5em minus 0.4em\relax IEEE, 2020, pp. 1746--1752.

\bibitem{LLM4Drive}
Z.~Yang, X.~Jia, H.~Li, and J.~Yan, ``{LLM4Drive}: A survey of large language models for autonomous driving,'' in \emph{NeurIPS 2024 Workshop on Open-World Agents}, 2024.

\bibitem{MLLMforAD}
C.~Cui, Y.~Ma, X.~Cao, W.~Ye, Y.~Zhou, K.~Liang, J.~Chen, J.~Lu, Z.~Yang, K.-D. Liao \emph{et~al.}, ``A survey on multimodal large language models for autonomous driving,'' in \emph{IEEE/CVF Conference on Computer Vision and Pattern Recognition}, 2024, pp. 958--979.

\bibitem{DriveLikeAHuman}
D.~Fu, X.~Li, L.~Wen, M.~Dou, P.~Cai, B.~Shi, and Y.~Qiao, ``Drive like a human: Rethinking autonomous driving with large language models,'' in \emph{IEEE/CVF Winter Conference on Applications of Computer Vision Workshops}.\hskip 1em plus 0.5em minus 0.4em\relax IEEE, 2024, pp. 910--919.

\bibitem{LMDrive}
H.~Shao, Y.~Hu, L.~Wang, G.~Song, S.~L. Waslander, Y.~Liu, and H.~Li, ``{LMDrive}: Closed-loop end-to-end driving with large language models,'' in \emph{IEEE/CVF Conference on Computer Vision and Pattern Recognition}, 2024, pp. 15\,120--15\,130.

\bibitem{GPTDriver}
J.~Mao, Y.~Qian, J.~Ye, H.~Zhao, and Y.~Wang, ``{GPT-Driver}: Learning to drive with gpt,'' in \emph{NeurIPS 2023 Foundation Models for Decision Making Workshop}, 2023.

\bibitem{DriveGPT4}
Z.~Xu, Y.~Zhang, E.~Xie, Z.~Zhao, Y.~Guo, K.-Y.~K. Wong, Z.~Li, and H.~Zhao, ``{DriveGPT4}: Interpretable end-to-end autonomous driving via large language model,'' \emph{IEEE Robotics and Automation Letters}, 2024.

\bibitem{Dilu}
L.~Wen, D.~Fu, X.~Li, X.~Cai, T.~MA, P.~Cai, M.~Dou, B.~Shi, L.~He, and Y.~Qiao, ``{DiLu}: A knowledge-driven approach to autonomous driving with large language models,'' in \emph{The Twelfth International Conference on Learning Representations}, 2024.

\bibitem{RRR}
C.~Cui, Y.~Ma, X.~Cao, W.~Ye, and Z.~Wang, ``Receive, reason, and react: Drive as you say, with large language models in autonomous vehicles,'' \emph{IEEE Intelligent Transportation Systems Magazine}, 2024.

\bibitem{VLM-MPC}
K.~Long, H.~Shi, J.~Liu, and X.~Li, ``{VLM-MPC}: Vision language foundation model -guided model predictive controller for autonomous driving,'' \emph{arXiv preprint arXiv:2408.04821}, 2024.

\bibitem{LLM4Robotics}
J.~Wang, E.~Shi, H.~Hu, C.~Ma, Y.~Liu, X.~Wang, Y.~Yao, X.~Liu, B.~Ge, and S.~Zhang, ``Large language models for robotics: Opportunities, challenges, and perspectives,'' \emph{Journal of Automation and Intelligence}, 2024.

\bibitem{CodeAsPolicies}
J.~Liang, W.~Huang, F.~Xia, P.~Xu, K.~Hausman, B.~Ichter, P.~Florence, and A.~Zeng, ``Code as policies: Language model programs for embodied control,'' in \emph{IEEE International Conference on Robotics and Automation}.\hskip 1em plus 0.5em minus 0.4em\relax IEEE, 2023, pp. 9493--9500.

\bibitem{LLM2Rewards}
W.~Yu, N.~Gileadi, C.~Fu, S.~Kirmani, K.-H. Lee, M.~G. Arenas, H.-T.~L. Chiang, T.~Erez, L.~Hasenclever, J.~Humplik \emph{et~al.}, ``Language to rewards for robotic skill synthesis,'' in \emph{Conference on Robot Learning}.\hskip 1em plus 0.5em minus 0.4em\relax PMLR, 2023, pp. 374--404.

\bibitem{SafetyTaskPlanningRobotics}
A.~A. Khan, M.~Andrev, M.~A. Murtaza, S.~Aguilera, R.~Zhang, J.~Ding, S.~Hutchinson, and A.~Anwar, ``Safety aware task planning via large language models in robotics,'' \emph{arXiv preprint arXiv:2503.15707}, 2025.

\bibitem{zeroCoT}
T.~Kojima, S.~S. Gu, M.~Reid, Y.~Matsuo, and Y.~Iwasawa, ``Large language models are zero-shot reasoners,'' \emph{Advances in neural information processing systems}, vol.~35, pp. 22\,199--22\,213, 2022.

\bibitem{FewshotLearner}
T.~Brown, B.~Mann, N.~Ryder, M.~Subbiah, J.~D. Kaplan, P.~Dhariwal, A.~Neelakantan, P.~Shyam, G.~Sastry, A.~Askell \emph{et~al.}, ``Language models are few-shot learners,'' \emph{Advances in neural information processing systems}, vol.~33, pp. 1877--1901, 2020.

\bibitem{IntermediateMPC}
K.~Honda, H.~Okuda, T.~Suzuki, and A.~Ito, ``Connection of nonlinear model predictive controllers for smooth task switching in autonomous driving,'' \emph{Asian Journal of Control}, vol.~25, no.~3, pp. 1805--1822, 2023.

\bibitem{HighwayEnv}
E.~Leurent, ``An environment for autonomous driving decision-making,'' \url{https://github.com/eleurent/highway-env}, 2018.

\bibitem{MPPI}
G.~Williams, P.~Drews, B.~Goldfain, J.~M. Rehg, and E.~A. Theodorou, ``Information-theoretic model predictive control: Theory and applications to autonomous driving,'' \emph{IEEE Transactions on Robotics}, vol.~34, no.~6, pp. 1603--1622, 2018.

\end{thebibliography}
